\documentclass[runningheads]{llncs}

\usepackage[T1]{fontenc}
\usepackage{url}
\usepackage{graphicx}
\usepackage{listings}
\usepackage[normalem]{ulem}
\usepackage{color} 
\usepackage{booktabs} 
\usepackage{amsmath} 

\definecolor{backcolour}{rgb}{0.95,0.95,0.92}
\definecolor{codegreen}{rgb}{0,0.6,0}
\definecolor{magenta}{rgb}{0.5,0,0.5}
\definecolor{codegray}{rgb}{0.5,0.5,0.5}
\definecolor{codepurple}{rgb}{0.58,0,0.82}

\lstdefinestyle{prompt_style}{
    backgroundcolor=\color{backcolour},   
    commentstyle=\color{codegreen},
    keywordstyle=\color{magenta},
    numberstyle=\tiny\color{codegray},
    stringstyle=\color{codepurple},
    basicstyle=\ttfamily\footnotesize,
    breakatwhitespace=true,         
    breaklines=true,                 
    captionpos=b,                    
    keepspaces=true,                 
    numbers=left,                    
    numbersep=5pt,                   
    showspaces=false,                
    showstringspaces=false,
    showtabs=false,                  
    tabsize=2
}

\begin{document}

\title{RED.AI Id-Pattern: First Results of Stone Deterioration Patterns with Multi-Agent Systems}

\titlerunning{RED.AI Id-Pattern: Stone Deterioration Patterns with MAS}

\author{ 
Daniele Corradetti\inst{1,2} \and José Delgado Rodrigues\inst{3}}

\authorrunning{Corradetti D. \& Delgado Rodrigues, J.}

\institute{STAP Reabilitação Estrutural, SA, Rua General Ferreira Martins 8 - 9B, Algés, 1495-137, Portugal \and Grupo de Fisica Matematica, Instituto Superior Tecnico, Av. Rovisco Pais, Lisboa, 1049-001, Portugal \and Consultant in Conservation of Cultural Heritage, Rua Cidade da Beira, 76-4E, Lisbon, 1800-070, Lisbon, Portugal}

\maketitle              

\begin{abstract}
The Id-Pattern system within the RED.AI project (Reabilitação Estrutural Digital através da AI) consists of an agentic system designed to assist in the identification of stone deterioration patterns. Traditional methodologies, based on direct observation by expert teams, are accurate but costly in terms of time and resources. The system developed here introduces and evaluates a multi-agent artificial intelligence (AI) system, designed to simulate collaboration between experts and automate the diagnosis of stone pathologies from visual evidence. The approach is based on a cognitive architecture that orchestrates a team of specialized AI agents which, in this specific case, are limited to five: a lithologist, a pathologist, an environmental expert, a conservator-restorer, and a diagnostic coordinator. To evaluate the system we selected 28 difficult images involving multiple deterioration patterns. Our first results showed a huge boost on all metrics of our system compared to the foundational model.
\end{abstract}
\section{Introduction}

In recent years, the evolution of AI technologies in the cultural heritage sector has marked significant progress: initially dominated by specific neural networks for image classification such as CNNs or U-Net for determining stone deterioration patterns or detecting cracks or erosion in historical structures \cite{trichopoulos2023,spennemann2023,vitaloni2025}. Subsequently, albeit still in a seminal way, the field has witnessed the rise of multimodal Large Language Models (LLMs), such as GPT-4o or similar, which simultaneously process visual and textual inputs. As we explained in what to our knowledge was the first work in this sector \cite{corradetti2025}, the use of LLMs enables comprehensive diagnosis, extending beyond mere visual categorization: LLMs can correlate degradation patterns observed in an image with textual descriptions from technical databases, generating structured outputs that include confidence levels, differential hypotheses, and even conservative recommendations \cite{corradetti2025}. A multimodal LLM can, for example, not only identify an area of "blackening" on a facade, but also contextualize it spatially, correlate it with textual data on black crusts or biocolonization, and finally formulate a diagnostic hypothesis based on inferential reasoning that integrates both sources of information.

However, the use of a monolithic LLM model presents intrinsic limitations: the potential generation of unverified information ("hallucinations"), the difficulty of ensuring adherence to specific terminological standards (e.g., ICOMOS-ISCS, 2008), and the "opaque" nature of its decision-making process. To overcome these criticalities, research is moving towards more structured cognitive architectures, such as agentic ones. An agentic architecture, in fact, decomposes a complex problem into specialized sub-tasks, assigning them to autonomous "agents" that collaborate to achieve a common objective. These architectures are known to be more robust and interpretable as well as generally more reliable and precise.

Anticipating the importance of an agentic approach through LLMs for the diagnosis of stone pathologies, in March 2023, STAP S.a., a Portuguese excellence in structural rehabilitation, initiated a self-funded project called RED.AI (Reabilitação Estrutural Digital através da AI). The Id-Pattern system (Identification of Stone Deterioration Patterns) that we present in this article falls within this project. It consists of a multi-agent architecture formed by a predetermined number of agents specialized in specific areas of cultural heritage conservation and restoration and coordinated by an orchestrator agent. This approach not only elevates objectivity and standardization, but represents an innovation compared to monolithic models, allowing a simulation of multidisciplinary debates that enriches the analysis with complementary perspectives \cite{kiourt2017,costantini2008}. In this article we present the first results of this work that show remarkable results, albeit simply preliminary. Indeed, as shown in Table 1, all metrics of the Id-Pattern system are significantly better than the foundational model. However, it is necessary to highlight how the sensitivity or recall of the system is particularly pronounced. The Id-Pattern system, in fact, doubles the sensitivity of the foundational model, without losing, but indeed also gaining in precision and robustness.

\section{Context and Technical Problems}

The world of multimodal LLMs has radically changed in the last two years, both in cost, inference capabilities, as well as in the possibility of being able to "reason" on images in addition to proceeding with immediate inference. Despite these profound changes in the technical landscape of LLMs, the structure of the RED.AI project has remained essentially unchanged. However, while having a clear understanding of the result to be obtained, it is necessary to specify that only in the last months of 2024 have the evolution of multimodal models and their inference cost become such as to justify a concrete realization of the project in its multi-agent version. For this reason, almost the entirety of 2024 was dedicated to the creation of a benchmark \cite{corradetti2025}  that would allow us to evaluate the reliability of the main foundational models (at the time consisting of OpenAI, Anthropic and Google), their strengths and the pathologies invisible to them or difficult to discriminate. Our preliminary analysis on the benchmark showed that, despite LLMs showing significant potential, fine-tuning and architectural enhancements were required to consistently and reliably identify certain patterns. Subsequent studies and internal attempts subsequently showed the insufficiency of simple prompt engineering techniques, few-shot learning, retrieval augmented generation (RAG) techniques and model fine-tuning. A detailed analysis of the reasons for the insufficiency of these techniques would require a systematic study and specific work, but for our understanding it is rooted in some fundamental elements:
\begin{itemize}
    \item The insufficiency of high-quality data specifically designed for stone pathology: foundational models have many visual data that have often been catalogued by non-specialists in stone heritage. The absence of large high-quality datasets is a deficiency that necessarily limits AI development in this sector.
    \item The fact that there exists an important differential between the meaning of terms in the context of stone pathology (for example "spalling and chipping") and the meaning of the same terms in a generic context can be the cause of  diagnostic errors in foundational models. This means that stone heritage datasets relating to terms that have a specific meaning in the context of stone pathology, but a common meaning outside of this, can be more easily contaminated by generic datasets compared to other sectors (for example advanced mathematics, physics and coding) where technical terms have no application outside the context.
    \item Absence of a specific procedure in analyzing images to identify what are the important elements from a diagnostic point of view and those that are superfluous or collateral.
    \item Problems dictated by the non-stability of diagnosis: foundational models are all probabilistic non-deterministic models and it is necessary to balance a game on temperature to be sure of having a model temperature high enough to be able to consider less common pathologies, but at the same time avoid introducing many false positives.
\end{itemize}
The Id-Pattern project presented here addresses all of the previous points. In this sense we note that if the problems at point 1 and point 2 can perhaps be reduced through appropriate RAG and fine-tuning techniques, and if point 3 can be partially resolved through prompt-engineering techniques, point 4 naturally suggests the construction of an agentic architecture, capable of presenting various solutions to the same problem, evaluating them, analyzing them and then synthesizing them after adequate filtering.

\section{The RED.AI Id-Pattern Agentic Structure}

\subsection{Multi-Agent Systems: Foundations and Distinction from Monolithic Models}

Agentic structures represent an advanced paradigm in AI that, while being known since the end of the last century \cite{costantini2008}, is powerfully emerging during 2025 \cite{mazzetto2024,kampelopoulos2025,zhang2025} due to its application to LLMs. In agentic structures, a system is composed of multiple autonomous "agents," namely software entities equipped with specific objectives, decision-making capabilities, and interaction mechanisms, which collaborate to solve complex tasks. Unlike foundation models, such as monolithic LLMs that generate outputs based on statistical patterns learned from vast datasets, agentic structures introduce a layer of cognitive orchestration: each agent can access external tools (for example, retrieval from knowledge bases), reason sequentially and iteratively, and dynamically adapt to context. The results differ in terms of robustness and interpretability: while a foundation model might produce a single response potentially subject to hallucinations or bias, an agentic structure generates outputs through multi-perspective consensus, reducing errors in complex diagnostic tasks and improving decision traceability.

\subsection{The Architecture of RED.AI Id-Pattern}

The RED.AI Id-Pattern system is a specific implementation of a multi-agent system designed for stone heritage diagnosis. Compared to other agentic structures, such as the Multi-Agent Debate (MAD) framework for general reasoning \cite{mazzetto2024}, REDAI emphasizes a domain-specific protocol and sequential phases inspired by human multidisciplinary workflows, improving consistency in diagnostic tasks. The architecture, schematically described in Figure 1, is based on a team of five specialized AI agents, each powered by a multimodal LLM model (in this case, OpenAI's o4-mini model) supported by a RAG process to anchor analyses to a vector knowledge base derived from specialized technical documents. The system comprises five specialized agents: (i) the lithologist, focused on lithological identification and texture/porosity analysis; (ii) the pathologist, expert in classifying degradation patterns such as black crusts or biocolonizations; (iii) the environmental expert, who evaluates exposure factors such as rain, wind, and pollution; (iv) the conservator-restorer, who identifies previous interventions and conservation status; (v) the diagnostic coordinator, who synthesizes contributions for an integrated output.

Each agent receives an "identity card" that defines its role, areas of competence, and some personality traits. This serves to constrain the LLM, forcing it to reason exclusively within a specific domain. Additionally, each agent follows a rigid base protocol that includes contextual analysis, systematic observation, compatibility assessment, and diagnostic synthesis. The RAG integrates a vector store with consolidated documents from expert-LLM interactions, ensuring adherence to standardized terminologies. Finally, the workflow proceeds as follows:
\begin{enumerate}
    \item \textbf{Individual Analysis Phase:} Each agent is presented with the image to analyze. The agent must formulate a preliminary diagnosis following its specialization and a rigid BASE\_PROTOCOL provided in the prompt and supported by RAG. This initial phase is comparable to the preparatory work that each expert performs individually before a coordination meeting. The objective is to generate a series of preliminary specialized analyses, based on objective data and a shared protocol. The result is not a simple interpretation of the image, but an augmented and anchored analysis. The agent is forced to formulate its diagnosis not only based on its pre-trained knowledge, but especially in light of the specific definitions and technical criteria provided by RAG. For example, it will not generically describe a "dark stain," but must classify it as diffuse dark biocolonization or black crust using the technical definitions provided. The output of this phase is a set of specialized, grounded, and traceable diagnoses.
    \item \textbf{Multidisciplinary Discussion Phase:} Individual analyses are shared among all agents. Each agent comments on colleagues' results, highlighting concordances and discordances, always from their specialized perspective. This phase simulates a technical coordination meeting. In turn, each agent "reads" the conclusions of its colleagues and is instructed to comment on them from their unique perspective. The output is not a single response, but a structured log of the discussion. In this phase, there is no real-time "negotiation," but rather a collection of opinions and counter-opinions that will serve as instructional material for the final phase. The system has now mapped areas of consensus and, more importantly, has isolated critical divergence points that require reasoned synthesis.
    \item \textbf{Consensus Phase:} The Diagnostic Coordinator receives the entire discussion log. Its task is to synthesize the different analyses, resolve divergences through logical reasoning, and formulate the Final Integrated Diagnosis, also indicating a confidence level (high, medium, low) based on the consistency of the emerged evidence.
\end{enumerate}

\begin{figure}
\includegraphics[width=\textwidth]{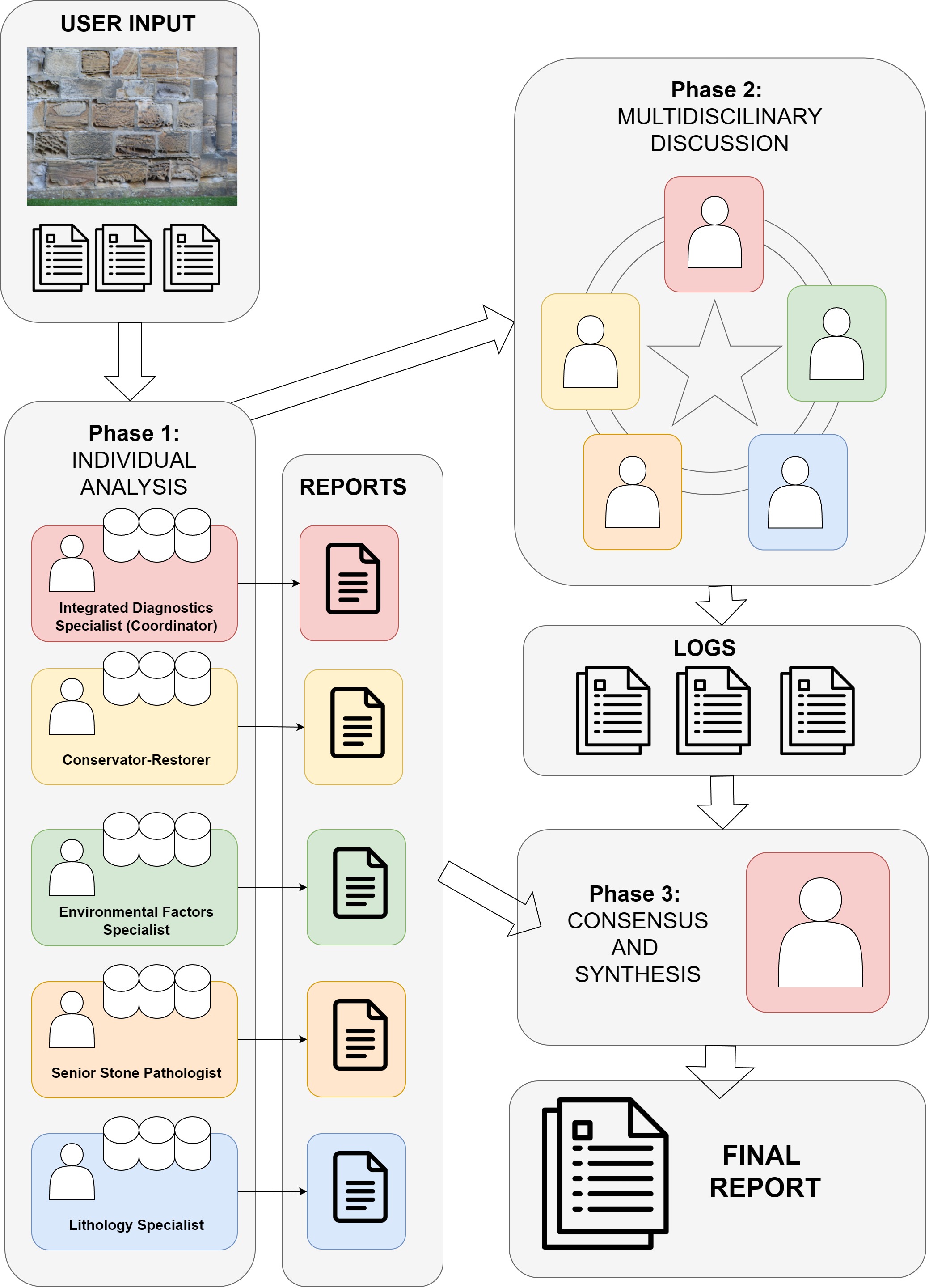}
\caption{Architecture of the RED.AI Id-Pattern multi-agent system. The flow shows parallel individual analysis, sequential discussion phase, and final synthesis coordinated by the diagnostic agent. The entire process is supported by a vector knowledge base interrogated through RAG.} \label{fig:1}
\end{figure}

\subsection{RAG Database Formation}

An important element for reducing the differential between common terminology and specialized terminology, as well as for reducing cases of hallucinations and misinterpretations, is provided by the RAG (Retrieval Augmented Generation) system with which each agent is equipped. Experimentally, we have found that, both for the volume of data collected and for the type of knowledge necessary to transmit to the models, the most effective solution for providing specialized instruction to individual agents is precisely that of RAG rather than through fine-tuning of the models. Although it is planned, in the most advanced version of the Id-Pattern system, that each agent be equipped with its own specific training, at the moment all agents possess a common knowledge block consisting of approximately 1M tokens. For the production of these texts to be useful, a long dialectical production process between one of the authors and the foundational model (in this case o4-mini) was necessary, aimed both at understanding the errors in the diagnostic process to which the model was prone and at understanding what is the best way to transmit and communicate corrections to the model itself. Through immediate feedback with the foundational model, it was possible to understand, through the analysis of numerous images, the most common errors and correct them with appropriate examples drawn from a proprietary database of thousands of images collected through decades of field experience.

\subsection{Description of the Base Protocol}

In addition to the set of specific knowledge provided to each agent through appropriate documents for RAG, individual agents have been provided with system instructions for image analysis defined as "Base Protocol". This protocol is not simply an instruction, but a true procedural reasoning schema imposed on each specialized agent during the individual analysis phase. The function of the protocol is twofold: on one hand, it standardizes the analysis method ensuring that all agents, regardless of their specialization, follow the same logical path ; on the other hand, it acts as a cognitive scaffolding that guides the LLM through a deliberative process that emulates that of a human expert, mitigating impulsive or superficial responses. The protocol is conceived to force an analysis that proceeds from general to particular, from observation to interpretation. Synthetically, without entering into the protocol details, we can say that this forces the agent not to immediately focus on detail, but to understand the context of the study object. The agent must identify the type of architectural element (e.g., vertical surface, decorative element), characterize its exposure conditions (e.g., exposure to rain, wind, pollution) and formulate a preliminary hypothesis on the nature of the lithology (e.g., carbonate or silicate). Subsequently, the protocol guides it to ideally subdivide the surface into homogeneous zones and to record visible phenomena (e.g., blackening, deposits, disaggregation, chromatic alterations), analyzing their distribution and continuity.Subsequently, the agent must verify the compatibility between the observed pattern, the exposure context and the presumed lithology, forcing the agent to validate or discard hypotheses based on logical rules. Finally, only after completing the previous steps, can the agent formulate a diagnostic synthesis.

\section{Evaluation of the Agentic Structure}

One of the main problems for such specialized domains is the shortage of benchmarks and tests that are correctly developed and at the same time on which LLMs have not been previously trained or contaminated. In our case, it was necessary to develop a specific test that would allow us to evaluate the gains of the Id-Pattern system compared to the foundational system on the points where the latter is weakest.

\subsection{Test Development}

To test the degree of effectiveness of the models, 27 images were selected, which had the condition of not having been used in the training previously carried out. They include open views of facades, small sections of walls, as well as close-up views of
deterioration patterns. Limestones and granites, as the main examples of carbonate and silicate rocks,
respectively, were the main lithotypes represented in the sampled images, but sandstones, marbles and volcanic rocks were also analyzed.

\subsection{Test Execution}

For each of the provided images, the same prompt was given to both the foundational model and the Id-Pattern system:
\begin{lstlisting}[style=prompt_style, frame=single, title={Prompt used for testing both the Id-Pattern system and the foundational model.}]
I would like you to analyze this image. I would like the output to be very concise and include the following points:

A. A brief description of the element and its context and, if feasible, the identification of the most likely lithological type.
B. Identification, in bullet form, of the deterioration patterns you can identify.

For better identification, indicate in each bullet where the respective pattern is present. At this stage, it is not necessary to present a discussion of the results or comments on the genesis of the patterns.
\end{lstlisting}
The foundational model (also used to power the Id-Pattern system) is o4-mini-high, a proprietary OpenAI model used in "effort=high" mode. Requests to OpenAI were made via API and for each question the Id-Pattern system took approximately 5 minutes and a cost of approximately \$0.30 total to provide a response, producing approximately 10k tokens of conversational logs between agents and <1k tokens of structured response according to the prompt instructions. The entire test of 27 images took approximately 2 hours and cost \$7.08.

\subsection{Test Correction}
The performance evaluation of both systems was conducted through a comparative analysis based on standard metrics in the field of artificial intelligence and machine learning. For each image, the list of deteriorations identified by each model was compared with a reference mapping ("ground truth") created by a human expert . From this comparison, the following quantitative metrics were extracted:
\begin{itemize}
    \item \textbf{True Positives (TP):} An instance in which the model correctly identified a deterioration pattern actually present in the image.
    \item \textbf{False Positives (FP):} An instance in which the model reported the presence of a deterioration that did not actually exist. This represents a "false alarm".
    \item \textbf{False Negatives (FN):} An instance in which the model failed to identify a deterioration that was present. This represents a "missed identification".
    \item \textbf{Ambiguous:} Cases in which the model's identification is partially correct or not sufficiently clear to be classified as TP or FP. These cases were counted but excluded from the calculation of main metrics.
\end{itemize}
Based on this data, two key performance indicators were calculated to evaluate the effectiveness of each system:
\begin{itemize}
    \item \textbf{Precision:} $Prec. = TP/(TP+FP)$ which measures the reliability of positive identifications made by the model. It is calculated as the ratio between true positives and the sum of true positives and false positives. High precision indicates that the model is reliable when it identifies a deterioration pattern.
    \item \textbf{Sensitivity (Recall):} $Rec = TP/(TP+FN)$ Measures the model's ability to find all deteriorations actually present. It is calculated as the ratio between true positives and the sum of true positives and false negatives.High recall indicates that the model is complete and does not overlook many deteriorations.
    \item \textbf{F1-Score:} $F1Score = 2 \times (Prec \times Rec)/(Prec + Rec)$. It is the harmonic mean of Precision and Recall. This metric provides a single summary value that balances both metrics, making it particularly useful for evaluating the overall performance of a system when one wants to avoid excessive polarization towards precision or recall.
\end{itemize}

\section{Results}
The results of the comparative test demonstrated a clear and consistent improvement in the performance of the Id-Pattern system compared to the foundational o4-mini-high model across all metrics.

\begin{table}
\centering
\caption{Comparative performance of the Foundational model and the Id-Pattern system.}\label{tab:1}
\begin{tabular}{@{}lcccccc@{}}
\toprule
System & TP & FP & FN & Precision & Recall & F1-score \\ \midrule
Foundational (o4-mini-high) & 73 & 51 & 126 &  58.9\%  &  36.7\% &  45.2\% \\
\textbf{Id-Pattern} & 149 & 51 & 65 & \textbf{74.5\%}  &  \textbf{69.6\%}  &  \textbf{72.0\%} \\ \bottomrule
\end{tabular}
\end{table}

Particularly relevant is the difference in sensitivity or Recall between the two systems, highlighting the agentic system's ability to drastically reduce omissions. The Id-Pattern system almost always doubled the recall value of the base model, thus proving to be much more sensitive. For example, in the analysis of an image with 10 actual deteriorations, the foundational model correctly identified only 4 (Recall of 40.0\%), while the Id-Pattern system identified 8 (Recall of 80.0\%). This trend was repeated in most tests: in another case, recall went from 20.0\% of the base model to 80.0\% of the Id-Pattern system, and in yet another from 37.5\% to 87.5\%. Conversely, the foundational model misses over 60\% of present pathologies (high FN), reducing its usefulness in contexts where completeness of diagnosis is crucial. From a practical standpoint, while the foundational model succeeded in only 5 cases in identifying at least 50\% of the pathologies present in the image, the Id-Pattern system identified at least 50\% of the pathologies in as many as 26 images.To the notable sensitivity is added a considerable increase in F1-Score, confirming the Id-Pattern system as a significantly more robust and balanced diagnostic system.

\section{Conclusions and Future Developments}
These first results indicate that the research course has proven suitable for improving the capabilities of AI models for the automatic diagnosis of stone deterioration patterns in built heritage. An initial training phase allowed for the development of basic knowledge for use in the agentic  pproach described here. In addition to the more accurate use of terminology, the results show a clear improvement in the Id-Pattern system capabilities in terms of diagnostic
effectiveness. This significant improvement, compared to the foundational model, is evident in all statistical parameters used to evaluate the results.
The next phase of the research will be dedicated to interpreting the reasons that may be at the origin of false positives, and the objective will also be to search for a new agent architecture that is expected to be more robust than the one documented here.

\begin{credits}
\subsubsection{\ackname} We would like to thank José Paulo Costa and STAP for promoting this research.  

\subsubsection{\discintname} The authors have no competing interests to declare that are relevant to the content of this article.
\end{credits}

\end{document}